\documentclass{article}
\usepackage{spconf,amsmath,graphicx}
\usepackage{cite}
\usepackage{amsmath,amssymb,amsfonts}
\usepackage{algorithmic}
\usepackage{graphicx}
\usepackage{textcomp}
\usepackage{xcolor}
\usepackage{booktabs}
\usepackage{hyperref}

\title{Uncovering Overconfident Failures in CXR Models via Augmentation-Sensitivity Risk Scoring}
\name{
\shortstack[c]{
Han-Jay Shu$^{1*}$, Wei-Ning Chiu$^{2*}$, Shun-Ting Chang$^{3*}$, Meng-Ping Huang$^{4}$, \\
{\textit{Takeshi Tohyama$^{5}$, Ahram Han$^{5}$, Po-Chih Kuo$^{1}$}}}\thanks{* These authors contributed equally to this work.}}
\address{$^{1}$National Tsing Hua University, Department of Computer Science\\
$^{2}$National Taiwan University, Department of Computer Science and Information Engineering\\
$^{3}$National Tsing Hua University, Department of Electrical Engineering and Computer Science\\
$^{4}$National Tsing Hua University, Department of Technology Management\\
$^{5}$Massachusetts Institute of Technology, Laboratory for Computational Physiology}

\begin{document}
\maketitle
%
%
\begin{abstract}
{
Deep learning models achieve strong performance in chest radiograph (CXR) interpretation, yet fairness and reliability concerns persist. Models often show uneven accuracy across patient subgroups, leading to hidden failures not reflected in aggregate metrics. Existing error detection approaches—based on confidence calibration or out-of-distribution (OOD) detection—struggle with subtle within-distribution errors, while image- and representation-level consistency-based methods remain underexplored in medical imaging. We propose an augmentation-sensitivity risk scoring (ASRS) framework to identify error-prone CXR cases. ASRS applies clinically plausible rotations ($\pm15^\circ$/$\pm30^\circ$) and measures embedding shifts with the RAD-DINO encoder. Sensitivity scores stratify samples into stability quartiles, where highly sensitive cases show substantially lower recall ($-0.2$ to $-0.3$) despite high AUROC and confidence. ASRS provides a label-free means for selective prediction and clinician review, improving fairness and safety in medical AI.
}
\end{abstract}
\section{Introduction}
Deep learning models achieve strong performance in medical imaging, including chest radiograph (CXR) interpretation\cite{0a}. Yet growing evidence shows concerns about fairness and reliability: models often perform unevenly across subgroups, such as sex or race, resulting in hidden failures that are overlooked by aggregate metrics\cite{001,002}. Subgroup analyses based on explicit labels are common, but less attention has been given to detecting error-prone samples directly from images or model representations\cite{gg}. This raises the risk of missing “hidden subgroups” driven by latent image characteristics or model behaviors not captured by annotations.

Most error detection methods rely on model confidence scores (e.g., softmax probability, entropy, margin)\cite{1a,1b}, but these are undermined by miscalibration—neural networks often produce overconfident yet wrong predictions\cite{2a,2b}. Out-of-distribution (OOD) detection methods (ODIN, Mahalanobis, energy scores)\cite{liang2017enhancing,3a} handle large distribution shifts but struggle with subtle within-distribution errors, such as acquisition or image-level differences\cite{4a}.

Consistency-based methods offer an alternative: perturbation stability. Test-time augmentation (TTA) measures prediction variability across augmented views\cite{5b}, and consistency regularization encourages robustness to perturbations\cite{6a}. While conceptually aligned with the intuition that unstable predictions signal higher risk, these methods have mainly been applied for robustness or semi-supervised learning, not error detection\cite{7a}.

To address this gap, we propose the Augmentation-Sensitivity Risk Scoring (ASRS) framework. ASRS evaluates how sensitive model representations are to small, clinically plausible perturbations. We show that highly sensitive cases correspond to lower diagnostic reliability despite appearing confident under standard metrics. This reveals a key failure mode—\textbf{overconfident but unstable} predictions—that existing approaches miss. ASRS thus provides a label-free tool for selective prediction and clinician review, supporting safer and fairer medical AI deployment.

\begin{figure*}[t]
    \centering
    \includegraphics[width=\textwidth]{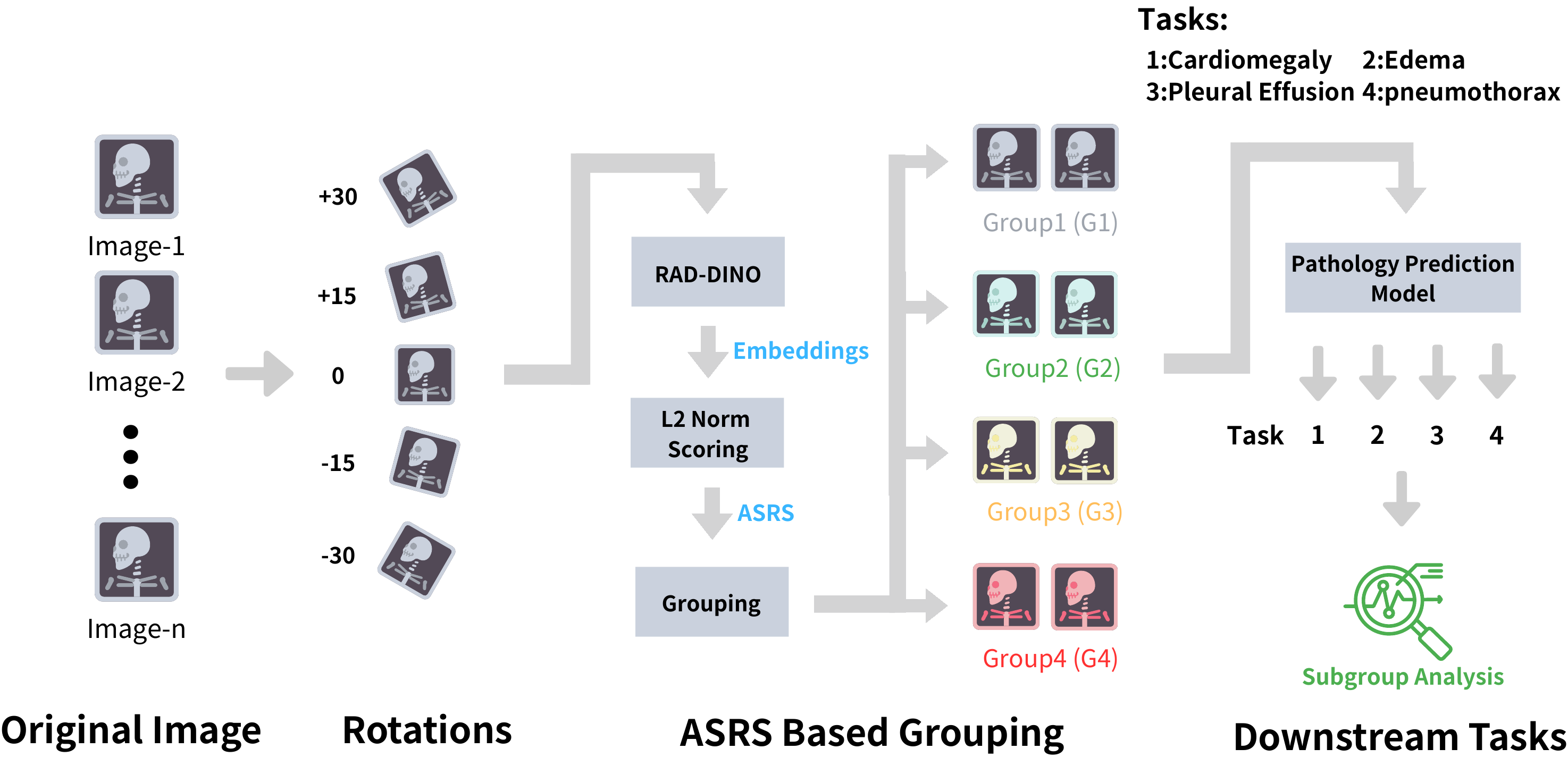}
    \caption{\textbf{Overview of the proposed methodology}, illustrating the pipeline from ASRS computation using RAD-DINO embeddings, validation-anchored grouping into G1--G4, downstream evaluation on four diagnostic tasks (pneumothorax, cardiomegaly, pleural effusion, edema) with multiple encoders (RAD-DINO, ResNet50, CXR-MAE), to stratified performance and confidence analysis.}
    \label{fig:overview}
\end{figure*}

\begin{table*}[t]
\caption{\textbf{Cohort characteristics of the MIMIC-CXR-JPG dataset (frontal PA/AP views).} Patient-level splits prevent leakage. Prevalence of diagnostic tasks is reported for the test set.}
\label{tab:cohort}
\centering
\scriptsize
\setlength{\tabcolsep}{3pt} 
\setlength{\tabcolsep}{3.5pt}
\begin{tabular}{lrrrrrrrrrrrr}
\toprule
\textbf{Split} & \textbf{Patients} & \textbf{Studies} & \textbf{Images} & \textbf{PA} & \textbf{AP} &
\textbf{Age (mean$\pm$SD)} & \textbf{Female \%} & \textbf{White \%} & \textbf{Black \%} & \textbf{Asian \%} & \textbf{Hisp./Lat. \%} & \textbf{Other/Unk. \%} \\
\midrule
Train & 30{,}238 & 117{,}686 & 131{,}283 & 47{,}650 & 83{,}633 &
$61.9 \pm 16.8$ & 45.8 & 66.6 & 16.3 & 3.0 & 5.4 & 8.8 \\
Val   & 10{,}079 & 38{,}846  & 43{,}335  & 15{,}619 & 27{,}716 &
$62.4 \pm 16.8$ & 46.0 & 67.1 & 15.4 & 3.7 & 5.4 & 8.4 \\
Test  & 10{,}081 & 38{,}444  & 42{,}918  & 15{,}815 & 27{,}103 &
$61.3 \pm 17.1$ & 45.9 & 65.5 & 15.9 & 3.9 & 5.8 & 9.0 \\
\midrule
\multicolumn{13}{l}{\textit{Test-set prevalence of diagnostic tasks (\%)}} \\
\multicolumn{2}{l}{Pneumothorax: 4.9} & \multicolumn{2}{l}{Cardiomegaly: 20.4} &
\multicolumn{2}{l}{Pleural Effusion: 23.8} & \multicolumn{2}{l}{Edema: 12.4} & & \\
\bottomrule
\end{tabular}
\end{table*}

\begin{table*}[t]
\caption{\textbf{Per-subgroup metrics for four diagnostic tasks across three architectures.} Each subgroup reports original Precision (Prec.), Recall (Rec.), AUROC (AUC), and Recall and AUROC under resampling (R) aligned to G4 prevalence.}
\label{tab:merged_metrics}
\centering
\setlength{\tabcolsep}{2.5pt}
\scriptsize
\begin{tabular}{llccccc ccccc ccccc ccc}
\toprule
Task & Model 
& \multicolumn{5}{c}{G1} & \multicolumn{5}{c}{G2} & \multicolumn{5}{c}{G3} & \multicolumn{3}{c}{G4} \\
& & Prec. & Rec. & AUC & Rec.(R) & AUC(R) 
   & Prec. & Rec. & AUC & Rec.(R) & AUC(R) 
   & Prec. & Rec. & AUC & Rec.(R) & AUC(R) 
   & Prec. & Rec. & AUC  \\
\midrule
Cardiomegaly 
& CXR-MAE   & 0.341 & 0.779 & 0.680 & 0.774 & 0.678 & 0.352 & 0.761 & 0.703 & 0.757 & 0.698 & 0.363 & 0.688 & 0.739 & 0.672 & 0.735 & 0.320 & \textbf{0.430} & \textbf{0.813} \\
& RAD-DINO  & 0.381 & 0.781 & 0.735 & 0.778 & 0.733 & 0.392 & 0.793 & 0.756 & 0.783 & 0.753 & 0.398 & 0.747 & 0.784 & 0.739 & 0.782 & 0.360 & \textbf{0.556} & \textbf{0.851}  \\
& ResNet50  & 0.325 & 0.775 & 0.650 & 0.768 & 0.644 & 0.335 & 0.754 & 0.671 & 0.750 & 0.665 & 0.336 & 0.663 & 0.705 & 0.657 & 0.707 & 0.299 & \textbf{0.420} & \textbf{0.797}  \\
\midrule
Edema 
& CXR-MAE   & 0.300 & 0.837 & 0.791 & 0.846 & 0.783 & 0.331 & 0.822 & 0.824 & 0.838 & 0.832 & 0.341 & 0.792 & 0.863 & 0.833 & 0.873 & 0.285 & \textbf{0.589} & \textbf{0.921} \\
& RAD-DINO  & 0.336 & 0.825 & 0.823 & 0.822 & 0.818 & 0.357 & 0.834 & 0.846 & 0.844 & 0.845 & 0.357 & 0.832 & 0.884 & 0.849 & 0.884 & 0.337 & \textbf{0.684} & \textbf{0.937} \\
& ResNet50  & 0.270 & 0.798 & 0.750 & 0.809 & 0.749 & 0.295 & 0.790 & 0.783 & 0.793 & 0.792 & 0.306 & 0.752 & 0.823 & 0.782 & 0.834 & 0.269 & \textbf{0.576} & \textbf{0.899} \\
\midrule
Pneumothorax 
& CXR-MAE   & 0.135 & 0.748 & 0.752 & 0.721 & 0.741 & 0.133 & 0.716 & 0.765 & 0.783 & 0.777 & 0.120 & 0.676 & 0.785 & 0.713 & 0.811 & 0.080 & \textbf{0.519} & \textbf{0.843}\\
& RAD-DINO  & 0.182 & 0.796 & 0.826 & 0.829 & 0.841 & 0.197 & 0.779 & 0.848 & 0.814 & 0.866 & 0.178 & 0.672 & 0.844 & 0.721 & 0.859 & 0.156 & \textbf{0.535} & \textbf{0.893} \\
& ResNet50  & 0.126 & 0.678 & 0.702 & 0.698 & 0.710 & 0.118 & 0.667 & 0.723 & 0.682 & 0.732 & 0.087 & 0.546 & 0.714 & 0.543 & 0.713 & 0.059 & \textbf{0.504} & \textbf{0.796} \\
\midrule
Pleural Effusion 
& CXR-MAE   & 0.494 & 0.831 & 0.768 & 0.834 & 0.768 & 0.509 & 0.809 & 0.803 & 0.805 & 0.799 & 0.509 & 0.744 & 0.846 & 0.750 & 0.839 & 0.455 & \textbf{0.553} & \textbf{0.885}  \\
& RAD-DINO  & 0.552 & 0.835 & 0.820 & 0.802 & 0.805 & 0.568 & 0.828 & 0.848 & 0.831 & 0.849 & 0.572 & 0.789 & 0.883 & 0.783 & 0.877 & 0.556 & \textbf{0.608} & \textbf{0.925}  \\
& ResNet50  & 0.458 & 0.796 & 0.717 & 0.793 & 0.709 & 0.466 & 0.781 & 0.756 & 0.779 & 0.750 & 0.456 & 0.722 & 0.800 & 0.719 & 0.795 & 0.333 & \textbf{0.485} & \textbf{0.844}  \\
\bottomrule
\end{tabular}
\end{table*}

\begin{table*}[t]
\caption{\textbf{Mean confidence by subgroup (overall, positive, negative) for each model and task.} Detailed explanation about the confidence calculation can be referred to Section ~\ref{sec:conf}.}
\label{tab:conf_by_group}
\centering
\setlength{\tabcolsep}{3pt}
\footnotesize
\begin{tabular}{llcccccccccccc}
\toprule
Task & Model & \multicolumn{3}{c}{G1} & \multicolumn{3}{c}{G2} & \multicolumn{3}{c}{G3} & \multicolumn{3}{c}{G4} \\
 &  & Ovr. & Pos. & Neg. & Ovr. & Pos. & Neg. & Ovr. & Pos. & Neg. & Ovr. & Pos. & Neg. \\
\midrule
Cardiomegaly & CXR-MAE & 0.674 & 0.684 & 0.670 & 0.675 & 0.678 & 0.675 & 0.692 & 0.662 & 0.701 & 0.784 & 0.671 & 0.795 \\
 & Rad-Dino & 0.693 & 0.710 & 0.688 & 0.708 & 0.715 & 0.705 & 0.728 & 0.708 & 0.733 & 0.828 & 0.713 & 0.839 \\
 & ResNet50 & 0.660 & 0.668 & 0.658 & 0.666 & 0.666 & 0.666 & 0.687 & 0.660 & 0.695 & 0.787 & 0.677 & 0.797 \\
\midrule
Edema & CXR-MAE & 0.755 & 0.772 & 0.751 & 0.762 & 0.768 & 0.761 & 0.791 & 0.754 & 0.796 & 0.892 & 0.736 & 0.897 \\
 & Rad-Dino & 0.770 & 0.780 & 0.767 & 0.789 & 0.795 & 0.788 & 0.819 & 0.796 & 0.823 & 0.915 & 0.784 & 0.920 \\
 & ResNet50 & 0.738 & 0.749 & 0.735 & 0.747 & 0.750 & 0.746 & 0.779 & 0.732 & 0.786 & 0.889 & 0.738 & 0.894 \\
\midrule
Pneumothorax & CXR-MAE & 0.726 & 0.764 & 0.723 & 0.730 & 0.743 & 0.729 & 0.759 & 0.736 & 0.760 & 0.833 & 0.712 & 0.834 \\
 & Rad-Dino & 0.770 & 0.819 & 0.766 & 0.789 & 0.827 & 0.787 & 0.820 & 0.795 & 0.821 & 0.900 & 0.781 & 0.902 \\
 & ResNet50 & 0.736 & 0.745 & 0.736 & 0.740 & 0.742 & 0.739 & 0.767 & 0.734 & 0.768 & 0.838 & 0.757 & 0.839 \\
\midrule
Pleural Effusion & CXR-MAE & 0.749 & 0.771 & 0.738 & 0.750 & 0.755 & 0.748 & 0.773 & 0.747 & 0.781 & 0.867 & 0.744 & 0.875 \\
 & Rad-Dino & 0.770 & 0.790 & 0.760 & 0.785 & 0.801 & 0.778 & 0.817 & 0.795 & 0.823 & 0.906 & 0.781 & 0.915 \\
 & ResNet50 & 0.724 & 0.737 & 0.718 & 0.728 & 0.732 & 0.727 & 0.750 & 0.722 & 0.758 & 0.851 & 0.729 & 0.859 \\
\bottomrule
\end{tabular}
\end{table*}

\section{Methodology}
Our methodology consists of three main components (Fig.~\ref{fig:overview}). 
First, we compute a label\textendash free \emph{augmentation\textendash sensitivity risk score (ASRS)} by applying small rotations to chest radiographs and measuring the representation shift using a contrastive encoder (RAD\textendash DINO\cite{RadDINO}). 
Second, quartile thresholds derived from the validation set are used to stratify the test set into four groups (G1--G4), representing increasing levels of sensitivity. 
Finally, we evaluate multiple model architectures on four canonical CXR diagnostic tasks to assess whether ASRS exposes hidden failure modes not captured by conventional confidence estimates.

\subsection{Augmentation--Sensitivity Risk Scoring (ASRS)}
The core concept of our approach is to quantify the stability of image representations under small, clinically plausible perturbations. We use RAD-DINO\cite{RadDINO}, a self-supervised contrastive learning model pre-trained on chest radiographs to produce consistent embeddings for augmented views of the same image. For each chest radiograph \( x \), we define a set of rotational transformations:
{\small
\[
\mathcal{T} = \{\mathrm{Rot}(-30^{\circ}), \mathrm{Rot}(-15^{\circ}), \mathrm{Rot}(+15^{\circ}), \mathrm{Rot}(+30^{\circ})\}.
\]}
Using RAD-DINO, we extract embeddings for the original image, \( z_0 = f(x) \), and for each perturbed image, \( z_t = f(t(x)) \), where \( t \in \mathcal{T} \) and \( f(\cdot) \) maps inputs to a 768-dimensional feature space (\(\mathbb{R}^{768}\)). The ASRS score is computed as the aggregated L2-norm of the embedding shifts:
\begin{equation}
s(x) = \sum_{t \in \mathcal{T}} \lVert z_t - z_0 \rVert_2.
\end{equation}
A higher \( s(x) \) indicates greater instability in the representation under rotational perturbations, potentially reflecting complex image features or artifacts that challenge downstream diagnostic tasks.

\subsection{Validation--Anchored Grouping}
\label{sec:grouping}
To enable label-free stratification of the test set, we derive quartile thresholds from the validation set. For each image \( x \) in the validation set, we compute the ASRS score \( s(x) \). Let \( \{s(x)\}_{x \in \mathrm{val}} \) denote the set of ASRS scores for all validation images. We calculate the 25th, 50th, and 75th percentiles, denoted as \( \tau_{25} \), \( \tau_{50} \), and \( \tau_{75} \), respectively. These thresholds are applied to the test set to define four groups:
\begin{align*}
&\mathrm{G1}: s(x) \leq \tau_{25}, \\
&\mathrm{G2}: \tau_{25} < s(x) \leq \tau_{50}, \\
&\mathrm{G3}: \tau_{50} < s(x) \leq \tau_{75}, \\
&\mathrm{G4}: s(x) > \tau_{75}.
\end{align*}
By construction, G1 contains images with the most stable representations under rotational perturbations, while G4 contains those with the least stable representations. This label-free, validation-anchored approach ensures reproducibility and prevents information leakage from the test set.


\subsection{Confidence Settings}
\label{sec:conf}
For each image \( x \) with predicted probability \( p(x) \), we define confidence measures to assess prediction certainty:
\begin{align*}
\mathrm{Conf}_{\mathrm{overall}}(x) &= \max\{p(x), 1 - p(x)\}, \\
\mathrm{Conf}_{\mathrm{pos}}(x) &= \mathrm{Conf}_{\mathrm{overall}}(x) \mid \mathrm{gt} = 1, \\
\mathrm{Conf}_{\mathrm{neg}}(x) &= \mathrm{Conf}_{\mathrm{overall}}(x) \mid \mathrm{gt} = 0,
\end{align*}
where \( \mathrm{gt} = 1 \) and \( \mathrm{gt} = 0 \) denote positive and negative ground-truth labels, respectively. These measures enable comparison of conventional confidence patterns with ASRS-defined instability across groups G1--G4 (Section~\ref{sec:grouping})

\section{Experiment and Discussion}
\subsection{Dataset and Experimental Setup}
\label{sec:dataset}
We utilize the MIMIC-CXR-JPG dataset\cite{Johnson2019MIMICCXRJPG}, restricting our analysis to frontal chest radiographs (posteroanterior [PA] and anteroposterior [AP] views). 
To prevent data leakage, we implement a patient-level split, dividing the dataset into training, validation, and test sets with proportions of 60\%, 20\%, and 20\%, respectively. This ensures that all images from a given patient are exclusively assigned to one split. Table~\ref{tab:cohort} summarizes the cohort statistics, including the number of patients, studies, and images per split, as well as the prevalence of diagnostic labels in the test set. 

We evaluate our proposed augmentation-sensitivity risk scoring (ASRS) framework across four chest radiography tasks—Cardiomegaly, Edema, Pneumothorax, and Pleural Effusion. For each task, we train and evaluate multiple model architectures, including RAD-DINO\cite{RadDINO}, ResNet50\cite{ResNet}, and CXR-MAE\cite{CXRMAE}, to assess the generalizability of the ASRS framework across diverse encoders.  Each model is trained on the training set, tuned on the validation set, and evaluated on the held-out test set, as described above.
We evaluate model performance using three primary metrics---Area Under the Receiver Operating Characteristic Curve (AUROC), precision, and recall (True Positive Rate, TPR). Unless otherwise specified, predictions are thresholded at 0.5 to compute precision and recall.
Performance metrics are reported separately for groups G1--G4 (Section~\ref{sec:grouping}), enabling analysis of how augmentation sensitivity correlates with diagnostic performance across model architectures. 
\subsection{Recall, AUROC, and Confidence Trends}
Stratifying cases into stability quartiles reveals a clear trend in table ~\ref{tab:merged_metrics}---recall decreases steadily from G1 to G4, with the most perturbation-sensitive cases (G4) showing a 0.25–0.30 recall deficit relative to more stable groups. Notably, even after resampling disease prevalence rates to match G4, the recall gap between G4 and other groups remains. This indicates that unstable cases are more likely to be missed, highlighting weaker diagnostic reliability.
In contrast, AUROC increases from G1 to G4, suggesting strong relative ranking ability within G4 despite poor recall at the global threshold. This apparent paradox reflects how AUROC captures ranking quality but not calibration or absolute sensitivity, giving an overly optimistic picture of unstable cases.
To validate this findings, we examine confidence to further clarify this mismatch---in table ~\ref{tab:conf_by_group}, G4 exhibits the highest mean confidence, particularly for negatives, while maintaining a similar mean confidence for positives compared to other groups, even though its recall is lowest. This reveals a critical failure mode—overconfident yet unstable predictions—that conventional confidence-based methods fail to expose.
\subsection{Intersection of ASRS and Demographics}
Demographic analysis shows shifts in age and racial composition across quartiles in table ~\ref{tab:asrs_demo}. These shifts suggest that augmentation sensitivity may partially reflect underlying demographic or acquisition heterogeneity\cite{aa}, but can not be fully explained by demographic prevalence
alone. Instead, ASRS captures an additional dimension of instability that intersects with—but is not reducible to—demographic variation. Reporting both performance and demographics by ASRS group promotes transparency and fairness in deployment.

\begin{table}[t]
\caption{Demographic characteristics of test set subgroups defined by ASRS quartiles.}
\label{tab:asrs_demo}
\centering
\footnotesize
\setlength{\tabcolsep}{5pt}
\begin{tabular}{lrrrrr}
\toprule
\textbf{Indicator} & \textbf{G1} & \textbf{G2} & \textbf{G3} & \textbf{G4} & \textbf{G4 vs.\ G1} \\
\midrule
N (images)              & 10,415 & 10,768 & 10,781 & 10,954 & -- \\
Age, mean (years)       & 64.83  & 64.11  & 62.38  & 53.90  & $-10.93$ \\
Female (\%)             & 47.45  & 43.95  & 43.76  & 48.40  & $+0.95$ \% \\
White (\%)              & 67.33  & 66.50  & 66.37  & 61.66  & $-5.67$ \% \\
Black (\%)              & 14.05  & 13.51  & 15.82  & 20.07  & $+6.02$ \% \\
Hispanic/Latino (\%)    &  4.04  &  4.81  &  5.29  &  8.97  & $+4.93$ \% \\
\bottomrule
\end{tabular}
\vspace{2pt}
\end{table}

\subsection{ASRS, Confidence, and Clinical Implications}
ASRS complements traditional confidence measures by capturing representation stability under small, clinically plausible perturbations. While confidence estimates (e.g., softmax probability, entropy) reflect proximity to the decision boundary, they fail to identify unstable cases. In the most perturbation-sensitive group (G4), we observe comparable positive confidence but elevated negative confidence, coupled with the lowest recall. This reveals a critical failure mode—high-confidence yet unstable predictions—that conventional metrics overlook.

Clinically, combining these two perspectives enables practical deployment strategies: auto-accept stable cases (G1/G2), flag unstable cases (G4) for review or adjusted thresholds, and abstain on low-confidence cases. In resource-limited settings, ASRS supports selective prediction by prioritizing the 20–25\% most unstable cases, improving safety without overwhelming clinician workload. By exposing hidden instabilities and guiding subgroup-specific calibration, ASRS provides a simple, label-free tool for safer and fairer deployment of medical AI.

\section{Conclusion}
We presented the Augmentation-Sensitivity Risk Scoring (ASRS) framework, a label-free approach that detects error-prone chest radiography cases by measuring representation stability under small, clinically plausible perturbations. Unlike confidence- or OOD-based methods, ASRS uncovers a critical failure mode—overconfident yet unstable predictions—that conventional metrics fail to reveal. 
Across multiple tasks and models, highly sensitive cases consistently showed lower recall despite high AUROC and confidence, confirming ASRS’s effectiveness in exposing hidden failures. By enabling stratification of high-risk cases, ASRS provides a practical basis for selective prediction, supporting safer and fairer deployment of medical AI.
\bibliographystyle{IEEEbib}
\bibliography{myrefs}

\end{document}